\documentclass[letterpaper, 10 pt, conference]{ieeeconf}  %

\usepackage[T1]{fontenc}
\usepackage{xcolor}
\usepackage{xspace}
\usepackage{graphicx}
\usepackage{amsmath}
\usepackage{amsfonts}
\usepackage{multirow}       %
\usepackage{makecell}       %
\usepackage{adjustbox}      %
\usepackage{booktabs}       %
\usepackage{colortbl}       %
\usepackage{xcolor}         %
\usepackage{array}          %
\usepackage{graphicx}
\usepackage{subcaption} %
\usepackage{url}

\IEEEoverridecommandlockouts                              %

\newcommand{\ourmethod}{GRASPLAT\xspace}
\newcommand{\ourmethodcomplete}{GRASPLAT\xspace}

\title{\LARGE \bf
\ourmethodcomplete: Enabling dexterous grasping through novel view synthesis
}

\author{Matteo Bortolon$^{1,2,4}$, Nuno Ferreira Duarte$^{3}$, 
Plinio Moreno$^{3}$, \\ Fabio Poiesi$^{1}$, Jos\'e Santos-Victor$^{3}$, Alessio {Del Bue}$^{2}$
\thanks{$^{1}$TeV, Fondazione Bruno Kessler, Trento, Italy, $^{2}$PAVIS, Fondazione Istituto Italiano di Tecnologia, Genoa, Italy, $^{3}$ISR, Instituto Superior T\'ecnico, Universidade de Lisboa, Lisbon, Portugal, $^{4}$University of Trento, Trento, IT. This project received support from the H2020 EU project RePAIR (Grant Agreement n$^\circ$ 	964854), the Center for Responsible AI, and MPR-2023-12- SACCCT Project 14935 AI.PackBot. This work was made during an internship of the first author in IST, Lisbon. Corresponding author \tt\small mbortolon@fbk.eu.}
}

\begin{document}

\graphicspath{ {./imgs/} }

\maketitle
\thispagestyle{empty}
\pagestyle{empty}

\begin{abstract}
Achieving dexterous robotic grasping with multi-fingered hands remains a significant challenge. 
While existing methods rely on complete 3D scans to predict grasp poses, these approaches face limitations due to the difficulty of acquiring high-quality 3D data in real-world scenarios. 
In this paper, we introduce \ourmethod, a novel grasping framework that leverages consistent 3D information while being trained solely on RGB images. 
Our key insight is that by synthesizing physically plausible images of a hand grasping an object, we can regress the corresponding hand joints for a successful grasp. 
To achieve this, we utilize 3D Gaussian Splatting to generate high-fidelity novel views of real hand-object interactions, enabling end-to-end training with RGB data. 
Unlike prior methods, our approach incorporates a photometric loss that refines grasp predictions by minimizing discrepancies between rendered and real images.
We conduct extensive experiments on both synthetic and real-world grasping datasets, demonstrating that \ourmethod improves grasp success rates up to 36.9\% over existing image-based methods.
Project page: \url{https://mbortolon97.github.io/grasplat/}
\end{abstract}

\section{Introduction}

The recent emergence of new humanoid platforms underscores the need for efficient solutions to enable grasping with multi-fingered robotic hands. 
While other types of grippers, such as parallel jaws, can achieve remarkable accuracy using extensive datasets \cite{fang2020graspnet}, they still struggle with handling everyday objects and a wide range of materials~\cite{kadalagere2023dexterousReview, ma2011dexterityAnalysis}.

To estimate a successful dexterous grasp, recent approaches transform the target object into a 3D model (such as a mesh or point cloud) and use a neural network to predict one or more potential grasping poses~\cite{newbury2023deep, xu2023unidexgrasp, xu2024dexterousGraspTransformer, mohammadi20233dsgrasp}. 
However, these methods present several challenges: 
i) obtaining complete 3D scans often requires multiple viewpoints, which may not be feasible in real-world testing scenarios~\cite{corona2020ganhand};
ii) generating accurate grasp poses can be time-consuming, requiring substantial 3D digitization effort, which limits the number of objects that can be processed; 
iii) depth sensors can struggle with transparent or reflective objects, and the time needed to acquire and process 3D models of novel objects can be prohibitive for real-time applications~\cite{karami20223glassSurface}. 
To mitigate these challenges, other approaches have explored solutions that can predict hand poses directly from RGB images~\cite{ye2023affordancediffusion, corona2020ganhand}. 
However, RGB-based methods often struggle with precision, as they rely solely on photometric information.
Without explicit 3D reasoning, object-hand collisions are more likely to occur.

In this work, we propose a dexterous grasping method, \ourmethod, that leverages 3D information while being trained solely on images.
The core idea is: \textit{if we can synthesize a plausible image of a hand grasping an object, reversing the process allows to optimize the hand parameters for a successful grasp.}
To synthesize the image, we utilize novel view synthesis techniques, specifically 3D Gaussian Splatting (3DGS)~\cite{kerbl20233Dgaussians}, which encapsulates a coherent 3D model capable of generating novel images of real hand-object grasps and their joint parameters. This step enables us to incorporate a photometric loss based on the image of the grasped object, hence allowing us to train our model end-to-end using only RGB data.

\ourmethod follows a similar initial step to other RGB-only approaches: a neural network predicts the 3D parameters of the human-like hand model, MANO \cite{romero2017mano}. 
However, unlike these methods, we leverage 3DGS's ability to generate high-fidelity scene representations that include both the object and the hand. 
The hand model is articulated and follows the MANO skeleton, linking the neural network's predictions to the MANO model. 
As a result, the appearance of the hand adapts based on the predictions.
Inspired by analysis-by-synthesis techniques~\cite{yen2020inerf}, we introduce an additional photometric loss term during training, which compares the predicted image with the ground truth. 
This allows the model to refine its grasp predictions by minimizing the discrepancy between the rendered and real images, thereby enhancing both accuracy and robustness.

We evaluate our method through extensive experiments on synthetic and real-world grasping datasets, demonstrating a significant improvement in grasp success rates over baseline approaches, up to 36.9 \% on YCB.
Our ablation studies highlight the role of the photometric loss term, confirming that it enhances the precision of grasp pose predictions.
By leveraging the strengths of 3DGS, our method provides an effective way to improve dexterous robotic grasping.

In summary, our contributions are as follows: 
\begin{itemize} 
    \item We introduce a novel view synthesis solution for grasping based on 3DGS, which refines hand pose predictions through a photometric loss. 
    \item We develop a training pipeline that, despite requiring only an RGB image during inference, leverages the 3D spatial information from the 3DGS model. 
    \item We create a synthetic dataset based on GraspXL~\cite{zhang2024graspxl}, enabling the training of 3DGS models. \end{itemize}

\begin{figure*}[t]
   \centering
   \includegraphics[width=0.92\textwidth]{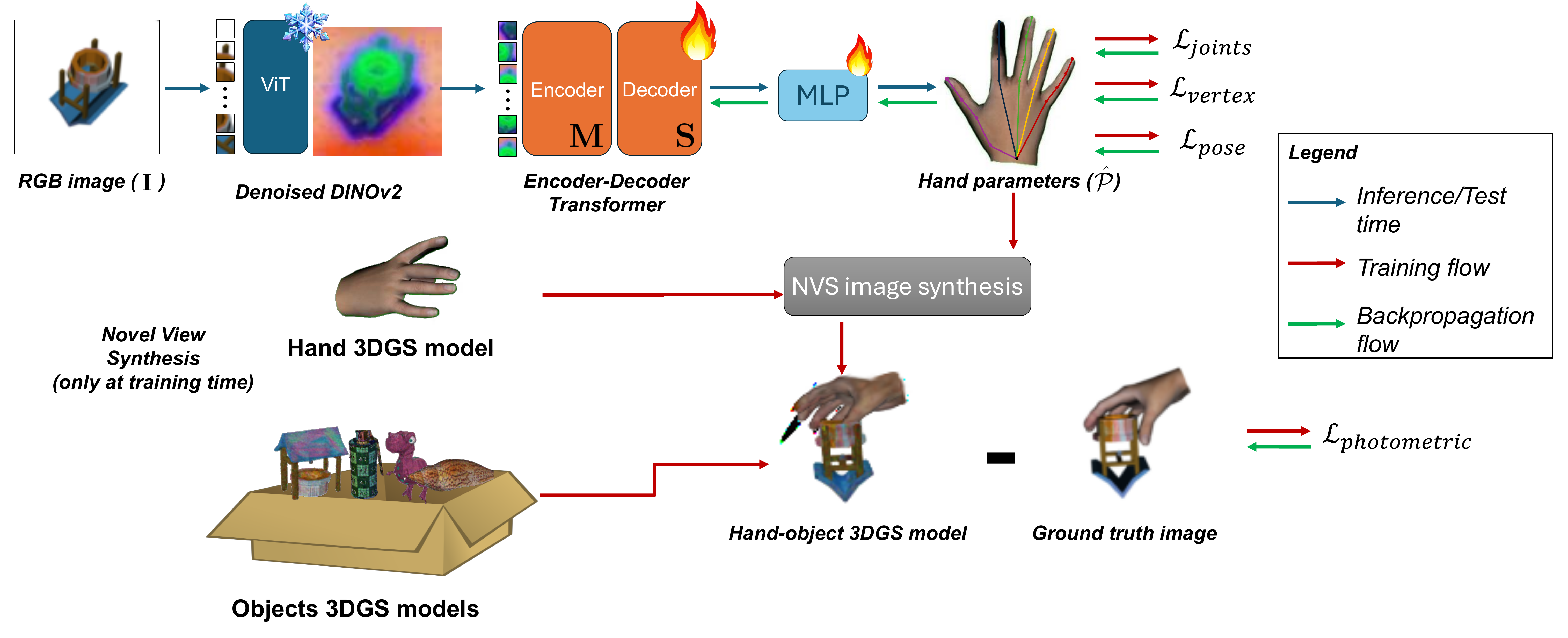}
   \caption{
      \ourmethod predicts 3D hand poses from a single RGB image using a frozen, denoised DINOv2 backbone.
      The extracted feature map is processed by a spatial transformer to estimate hand parameters, which are then refined by an MLP.
      During training, a feedback mechanism based on 3D Gaussian Splatting (3DGS) refines predictions by comparing rendered and real images.
      To accelerate training, 3DGS models are precomputed before training and then cached.
      The rendered scene includes both the object and the articulated hand, posed according to the predicted parameters.
   }
   \label{fig:diagram}
   \vspace{-5mm}
\end{figure*}

\section{Related work}

We categorize existing approaches as either geometric, which relies on 3D models of the object to predict the grasp pose, and RGB-based, which uses an image as input.

\vspace{1mm}
\noindent\textit{Geometric Methods.}
UniDexGrasp~\cite{xu2023unidexgrasp} follows a two-stage pipeline for dexterous grasping. Firstly, a grasp rotation is sampled, and a normalizing flow neural network predicts the translation and joint angles. 
These predictions are then refined using an ideal contact map. 
Secondly, a reinforcement learning-based grasping policy further refines the grasp.
The Dexterous Grasp Transformer~\cite{xu2024dexterousGraspTransformer} encodes the object’s point cloud using a PointNet++ network, and a transformer generates a set of grasp poses. 
The method initially learns coarse grasp poses, and then refines them. 
At test time, an additional refinement step further improves the results.
Similarly, \ourmethod leverages transformers to correlate different parts of the object. However, unlike these methods, \ourmethod assumes the input is an image rather than a complete point cloud of the object.

\vspace{1mm}
\noindent\textit{RGB Methods.}
Few methods have explored predicting grasps from a single RGB image of an object~\cite{corona2020ganhand, ye2023affordancediffusion}. GanHand~\cite{corona2020ganhand} follows a three-stage pipeline. 
First, it estimates the object's shape or pose, then predicts a grasp using a CNN, and lastly refines the hand parameters based on the reconstructed object. 
While our method also predicts hand parameters directly, as GanHand does, it does not require the object reconstruction step.
Affordance Diffusion~\cite{ye2023affordancediffusion} extends diffusion models to human grasp prediction by predicting a hand-object interaction layout, and then synthesizes realistic hand grasps. 
Then, it uses a pre-trained hand pose estimation network to estimate hand parameters. Differently, \ourmethod directly visualizes the predicted pose during training by leveraging the deterministic properties of 3DGS as opposed to relying on a probabilistic diffusion model.

\noindent\textit{Parametric MANO model.}
The MANO model~\cite{romero2017mano} parametrizes human hand deformation from an average mesh template based on pose parameters $\hat{\mathcal{P}} = \{\hat{\mathbf{Y}}, \hat{\mathbf{t}}, \hat{\mathbf{h}}\}$, where $\hat{\mathbf{Y}} \in \mathbb{R}^{3 \times 3}$ is the global rotation of the hand, $\hat{\mathbf{t}} \in \mathbb{R}^{3}$ is the translation, and $\hat{\mathbf{h}} \in \mathbb{R}^{21 \times 3 \times 3}$ represents joint rotations, relative to the parent in the kinematic chain, with the joint length fixed.

\section{\ourmethod}

\subsection{Overview}

Our approach estimates a 3D hand pose (in the MANO convention), denoted as $\hat{\mathcal{P}}$, from an RGB image $\mathbf{I}$.
Fig.~\ref{fig:diagram} shows the proposed approach diagram.
\ourmethod uses a Vision Transformer (ViT) to process $\mathbf{I}$ and extract a feature map that represents object shape clues.
This clues are fed into a transformer-based encoder-decoder architecture that predicts the hand pose $\hat{\mathcal{P}}$ (Sec.~\ref{sec:neuralnetwork}). 
During training, our model learns to output the hand parameters while being constrained by additional loss terms (Sec.~\ref{sec:hand_losses}).
We introduce a module that combines a 3DGS model of the object with an articulated hand model in a differentiable manner. 
The hand model is controlled by parameters predicted by a neural network. 
The combined model is then rendered from a randomly chosen viewpoint during training.
We compare this rendered view with a real scene image taken from the same viewpoint, where the hand has grasped the object. 
The comparison uses photometric error as the evaluation metric. 
The differentiability of 3DGS enables us to backpropagate the error to the neural network’s predictions (Sec.~\ref{sec:3dgs_phot_loss}).
Given the lack of a dataset combining grasping and 3DGS object models, we create our own dataset based on GraspXL~\cite{zhang2024graspxl} (Sec.~\ref{sec:visual_graspxl}).
At test time, the network generates the hand parameters for the multi-fingered robotic hand.

\subsection{Architecture}\label{sec:neuralnetwork}

We compute the feature map $\mathbf{F}$ from the image $\mathbf{I}$ using a pretrained DINOv2 Vision Transformer (ViT-B/14)~\cite{oquab2023dinov2}. 
We use a variant of DINOv2, called the Denoised Vision Transformer (DVT)~\cite{yang2024dvt}. 
Specifically, DVT adds an extra layer after the DINOv2 output to mitigate noise caused by the positional encoding inherent to transformers.

Since dexterous grasping requires high precision, we enhance the last layers of our neural network. 
We use an input image size of $512 \times 512$ as opposed to the typical $224 \times 224$. 
The image is then patchified, with each patch being processed individually. 
While we retain the standard patch size of $14 \times 14$, we reduce the stride from $14 \times 14$ to $7 \times 7$ to increase the resolution of the feature map. 
This modification expands the feature map $\mathbf{F}$ from its original size of $C \times 16 \times 16$ to $\mathbf{F} \in \mathbb{R}^{C \times 73 \times 73}$, where $C$ denotes the number of channels. 
We denote the width and height of the feature map as $E = 73$ and $T = 73$, respectively.

To enhance spatial awareness, we concatenate a 2D positional encoding to each output feature~\cite{carion2020detr}:
\begin{equation}
PE(x, y) = \begin{bmatrix} 
\sin\left(\frac{x}{\tau^{\frac{2i}{D}}}\right) \\
\cos\left(\frac{x}{\tau^{\frac{2i+1}{D}}}\right) \\
\end{bmatrix}, \quad \forall i \in [0, D/2),
\end{equation}
where $x$ and $y$ are the 2D coordinates, $D=\lfloor C / 2 \rfloor$ is positional encoding dimension, $i$ indexes the encoding dimension, and $\tau$=$10000$ is the temperature parameter.

We then fed these features into an encoder-decoder transformer~\cite{vaswani2017attention} to predict hand coordinates.
Due to its structure, the encoder-decoder output a spatial map $\mathbf{M} \in \mathbb{R}^{21 \times E \times T}$ and a global representation $\mathbf{S} \in \mathbb{R}^{C}$.
The representation $\mathbf{M}$ provides 2D hand joint positions, as a heatmap~\cite{fu2023deformer}.
Instead, $\mathbf{S}$ encodes the overall hand rotation and translation in 3D.

The heatmap is computed as $\mathbf{H}_j = \text{softmax}(\beta_j \mathbf{M}_j)$, where $j$ indexes the joint, and $\beta_j$ is a learned parameter that adjusts the weighting of each channel.
The final 2D coordinates for each joint are obtained via:
\begin{equation}
\mathbf{p}_j = \sum_{u}^{T} \sum_{v}^{E} \mathbf{H}_{j,u,v} \begin{bmatrix} u \\ v \end{bmatrix}.
\end{equation}

Each joint position $\mathbf{p}_j$ is concatenated with $\mathbf{S}$ then processed by a 4-layer MLP to estimate the wrist rotation $\hat{\mathbf{Y}}$, per-joint 3D rotation matrix $\hat{\mathbf{h}}$, and the wrist translation $\hat{\mathbf{t}}_n$. 
To ensure training stability, we normalize the translation with a mean of 0 and a standard deviation of 1. 
The final translation is then de-normalized as $\hat{\mathbf{t}} = \hat{\mathbf{t}}_n \mathbf{\sigma} + \mathbf{\mu}$, where the standard deviation $\mathbf{\sigma} \in \mathbb{R}^3$ and the mean $\mathbf{\mu} \in \mathbb{R}^3$ are computed from the training dataset.

\subsection{Loss functions for hand parameters prediction}\label{sec:hand_losses}
During training, multiple loss functions are used to supervise hand prediction.
Sec.~\ref{sec:datasets} provides details on how to efficiently generate the ground truth data for training \ourmethod.
Some loss functions are based on the vertex $\mathbf{V} \in \mathbb{R}^{768 \times 3}$ and joints $\mathbf{J} \in \mathbb{R}^{21 \times 3}$ computed using the MANO~\cite{romero2017mano} parametric model. 
In particular, a 3D vertex loss enforces geometry consistency between predicted $\hat{\mathbf{V}}$ and ground-truth mesh vertices $\mathbf{V}$ of MANO:
\begin{equation}
\mathcal{L}_{verts} = \lambda_{verts} \cdot \frac{1}{768} \sum_{i=1}^{768} \|\hat{\mathbf{V}}_i - \mathbf{V}_i\|_2^2,
\end{equation}
where $\lambda_{verts} = 1\times 10^{-4}$ is a balance factor.
The 3D joint loss aligns skeletal joint positions using:
\begin{equation}
\mathcal{L}_{joints} = \lambda_{joints} \cdot \frac{1}{21} \sum_{i=1}^{21} \|\hat{\mathbf{J}}_i - \mathbf{J}_i\|_2^2,
\end{equation}
where $\lambda_{joints} = 1\times 10^{-4}$.
Since a kinematic chain generate the vertex and joints, these losses also improve hand consistency.
To improve training time, we also directly supervise the MANO parameters including a loss term for rotation:
\begin{equation}
\mathcal{L}_{pose} = \lambda_{pose} \cdot \frac{1}{21} \sum_{i=1}^{21} \|\hat{\mathbf{Y}}_i - \mathbf{Y}_i\|_2^2,
\end{equation}
where $\lambda_{pose}=10.0$ balances the rotation component. 
We also include a term to supervise translation:
\begin{equation}
\mathcal{L}_{transl} = \lambda_{transl} \cdot \frac{1}{21} \sum_{i=1}^{21} \|\hat{\mathbf{t}}_i - \mathbf{t}_i\|_2^2,
\end{equation}
where $\lambda_{transl}=10.0$ weights the translation component. 
As in \cite{fu2023deformer}, a pose regularization term discourages unnatural hand configurations by penalizing large rotations:
\begin{equation}
\mathcal{L}_{reg} = \lambda_{reg} \cdot \frac{1}{21} \sum_{i=1}^{21} \| \mathbf{I}_i - \hat{\mathbf{Y}}_i \|_2^2,
\end{equation}
where $\mathbf{I} \in \mathbb{R}^{3 \times 3}$ is the identity matrix, and $\lambda_{reg}=1.0$. 
Lastly, the MANO part of the loss function is computed as:
\begin{equation}
\mathcal{L}_{mano} = \mathcal{L}_{verts} + \mathcal{L}_{joints} + \mathcal{L}_{pose} + \mathcal{L}_{transl} + \mathcal{L}_{reg}.
\end{equation}

\subsection{NVS photometric loss for grasping}\label{sec:3dgs_phot_loss}

\ourmethod uses a parameterized NVS model for articulated hands based on 3DGS that deforms according to the MANO mesh grid. 
We deform the 3DGS model using the mesh structure from the MANO parametric model. 
Such NVS model makes it possible to link the hands parameters to image generation and thus simplifying our training strategy using only images of hand-objects interactions.

\vspace{1mm}
\noindent\textit{3DGS preliminaries.}
The primary objective of 3DGS is to synthesize novel views of a scene using a set of 3D Gaussians, denoted as $\mathcal{Q} = \{\mathbf{Q}_i\}_{i=1}^{K}$. 
Each Gaussian is characterized by its position $\mathbf{x} \in \mathbb{R}^{3}$, orientation $\mathbf{R} \in SO(3)$, scale represented as a diagonal matrix $\mathbf{S} \in \mathbb{R}^{3 \times 3}$, and a color. 
A 3DGS model is created from a set of input images from different viewpoints.
The 3DGS model is optimized through an iterative process in which the Gaussians $\mathbf{Q}$ are projected onto the image plane and transformed into an image using the differentiable rasterization function $\phi$~\cite{kerbl20233Dgaussians}. 
The resulting image is then compared with the original image by minimizing a photometric error, and the parameters are optimized via backpropagation.
For more details on the process, please refer to~\cite{kerbl20233Dgaussians}.

\vspace{1mm}
\noindent\textit{Hand 3DGS model for \ourmethod.}
A standard 3DGS approach can synthesize images for rigid scenes only, so we upgraded 3DGS to generate images of articulated hands for grasp predictions. 
Specifically, any update to the MANO parameters $\hat{\mathcal{P}} = {\hat{\mathbf{Y}}, \hat{\mathbf{t}}, \hat{\mathbf{h}}}$ results in a corresponding deformation of the 3DGS model. 
To achieve this, we associate each Gaussian $\mathbf{Q}_i$ to the nearest face $w$ of the MANO mesh. 
Consequently, the transformation of each Gaussian is dependent on its corresponding face of the deforming mesh:
\begin{equation}
\hat{\mathbf{x}}_i = \mathbf{T}_w + k_w \mathbf{G}_w \mathbf{x}_i,
\end{equation}
\begin{equation}
\hat{\mathbf{R}}_i^{'} = \mathbf{R}_w \mathbf{R}_i,
\end{equation}
\begin{equation}
\hat{\mathbf{S}}_i^{'} = k_w \mathbf{S}_i^0,
\end{equation}
where $\mathbf{x}_i$, $\mathbf{R}_i$, and $\mathbf{S}_i$ represent the $i$th Gaussian position, rotation, and scaling respectively in the rest pose of MANO.
The transformation parameters $\mathbf{T}_w \in \mathbb{R}^3$ (translation), $\mathbf{G}_w \in SO(3)$ (rotation matrix), and $k_w \in \mathbb{R}^3$ (scaling factor) are derived from the MANO parameters for face $w$.
The translation parameters for each face $w$ is calculated as follows:
\begin{equation}
\mathbf{T}_w = \frac{1}{3}\sum_{i=0}^2 \mathbf{V}_{w_i},
\end{equation}
where $\mathbf{V}_{w_0}, \mathbf{V}_{w_1}, \mathbf{V}_{w_2}$ are the vertices of face $w$ in the MANO mesh.
The translation $\mathbf{T}_w$ is computed as the centroid of these vertices.
The rotation $\mathbf{R}_w$ derives from an orthogonal basis $\langle \cdot \rangle$ from the face's edge directions:
\begin{equation}
\begin{split}
\mathbf{R}_w = \langle \mathbf{a}_1, \mathbf{a}_2, \mathbf{a}_3 \rangle, \quad \text{where} \quad 
\mathbf{a}_1 &= \mathbf{V}_{w_2} - \mathbf{V}_{w_1}, \\
\mathbf{a}_2 &= \mathbf{a}_1 \times (\mathbf{V}_{w_0} - \mathbf{V}_{w_1}), \\
\mathbf{a}_3 &= \mathbf{a}_1 \times \mathbf{a}_2.
\end{split}
\end{equation}
The scaling factor $k_w$ is given by:
\begin{equation}
k_w = \frac{1}{2} \left ( \| \mathbf{a}_1 \|_2 + \frac{\| \mathbf{a}_2 \|_2}{\|\mathbf{V}_{w_0} - \mathbf{V}_{w_2}\|_2} \right ),
\end{equation}
and determined by averaging the face area scaling factor $\| \mathbf{a}_2 \|_2$ and the edge length variations $\| \mathbf{a}_1 \|_2$, ensuring robustness against significant distortions.

During training, we combine the hand and object 3DGS models using a simple concatenation $\mathcal{Q}_{scene} = \mathcal{Q}_{hand} \cup \mathcal{Q}_{object}$, where $\mathcal{Q}_{hand}$ is given by  the articulated hand 3DGS model, and $\mathcal{Q}_{object}$ represents the rigid object 3DGS model.
The combined 3DGS model $\mathcal{Q}_{scene}$ is rendered using the rendering function $\phi$, producing the final image $\hat{\mathbf{I}}$.
To prevent convergence to local minima, the viewpoint is randomly selected from a predefined set of viewpoints around the object.
The corresponding target image is denoted as $\mathbf{I}_{tgt}$ and
the photometric loss is computed as:
\begin{equation}\label{eqn:loss_photometric}
    \mathcal{L}_{img} = \lambda_1 \| \hat{\mathbf{I}} - \mathbf{I}_{gt} \|_1 + (1 - \lambda_1) (1 - \text{SSIM}(\hat{\mathbf{I}}, \mathbf{I}_{tgt})),
\end{equation}
where $\lambda_1 = 0.8$ is a balancing factor between the L1 term, which ensures convergence due to its convex nature, and the SSIM metric~\cite{wang2004ssim}, which, while non-convex, more closely aligns with human visual perception.
This formulation enables backpropagation of the visualization error into the pose predictions, resulting in the loss:
\begin{equation}\label{eqn:loss}
\mathcal{L} = \alpha \mathcal{L}_{mano} + (1 - \alpha) \mathcal{L}_{img}   
\end{equation}
where $\alpha = 0.5 ( 1 + cos( \pi o_{curr}/ o_{last}))$ gradually activates the photometric error, and $o_{curr}$, $o_{last}$ are the current and last epoch.

\subsection{\ourmethod dataset}\label{sec:visual_graspxl}
To train our model, we need to synthesize images of hand-object grasps from multiple viewpoints. 
We build upon the GraspXL~\cite{zhang2024graspxl} dataset, which contains 3D grasping sequences for 5,849 objects across various hand models. 
We extend GraspXL in several ways. 
First, we introduce a NVS pipeline that can generate images from arbitrary camera viewpoints. 
Our focus is on grasp sequences using the MANO hand model. 
Since GraspXL provides only geometric data without the textures needed for image synthesis, we augment the object generation process with  Objaverse~\cite{deitke2023objaverse}, which allows us to significantly increase the number of textured objects. 
Similarly, the appearance of the MANO hand is reconstructed using the DART texture model~\cite{gao2022dart}.
To train the 3DGS model, we render 200 images per object from multiple viewpoints using Vedo~\cite{musy2024vedo}. 
Each object undergoes five grasping attempts with different orientations, and we simulate the sequence until the object moves. 
Given \ourmethod's focus on humanoid robots, the hand begins from the bottom right of the camera view. 
In total, the dataset consists of over 4.5 million images, and it will be made publicly available upon paper acceptance. 
Qualitative examples of the synthesized data are shown in Fig.~\ref{fig:qualitative_results}.

\section{Experiments}

\subsection{Experimental setup}\label{sec:datasets}

\begin{table*}[t]
    \tabcolsep 3pt
	\caption{We report the grasp accuracy, mean joint error, relative rotation error (RRE), translation error, where applicable, along with inference time.
    Best-performing results are highlighted in \textbf{bold}.
    }
    \label{tab:comparison_results}
    \vspace{-2mm}
    \centering
    \resizebox{0.90\linewidth}{!}{%
	\begin{tabular}{l|cccc|c|c}
    \toprule
    & \multicolumn{4}{c|}{\ourmethod dataset} & YCB~\cite{Calli2015ycb} & \\
    Method & Grasp accuracy [\%] & Mean Joint Error [cm] & RRE [\textdegree] & Translation Error [cm] & Grasp accuracy [\%] & Time [s] \\
    \midrule
    GANHand~\cite{corona2020ganhand} & 37.1 & 2.22 & 16.23 & 7.38 & 26.3 & \textbf{0.26} \\
    Affordance Diffusion~\cite{ye2023affordancediffusion} & 34.2 & 2.11 & 16.03 & 7.00 & 0.0 & 0.83 \\
    \ourmethod (ours) & \textbf{61.4} & \textbf{1.58} & \textbf{15.15} & \textbf{5.23} & \textbf{63.2} & 0.38 \\
    \midrule
    $\Delta$ (best - second best) & {\color{teal}+24.3} & {\color{teal}+0.53} & {\color{teal}+0.88} & {\color{teal}+1.77} & {\color{teal}+36.9} & {\color{purple}-0.12} \\
    \midrule
    \midrule
    \ourmethod (224$\times$224 image resolution) & 60.0 & 1.59 & 15.43 & 5.36 & 52.6 & 0.13 \\
    \ourmethod (stride 14) & 58.5 & 1.59 & 15.19 & 5.26 & 47.4 & 0.12 \\
    \ourmethod (w/o $\mathcal{L}_{pose}$ \& $\mathcal{L}_{transl}$) & 60.1 & 1.66 & 15.18 & 5.60 & 62.9 & 0.38 \\
    \bottomrule
	\end{tabular}
 }
 \vspace{-2mm}
\end{table*}

\begin{figure*}[htb]
    \centering
    \begin{subfigure}[b]{\textwidth}
      \centering
      \renewcommand{\arraystretch}{1.2} %
      \begin{tabular}{ccccc}
        GANHand \cite{corona2020ganhand} & Affordance Diffusion \cite{ye2023affordancediffusion} & Ours (Simulation) & Ours (3DGS) & GT \\
        \includegraphics[width=0.156\textwidth]{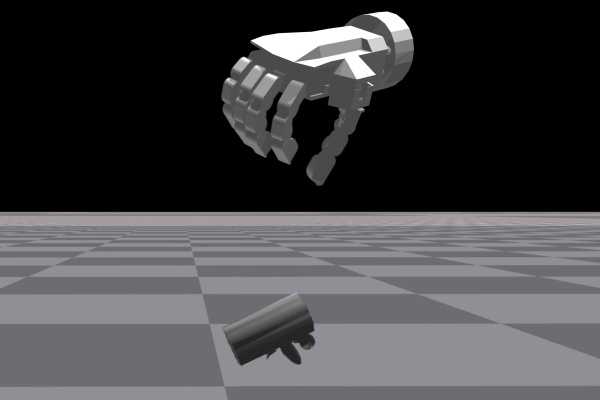} &
        \includegraphics[width=0.156\textwidth]{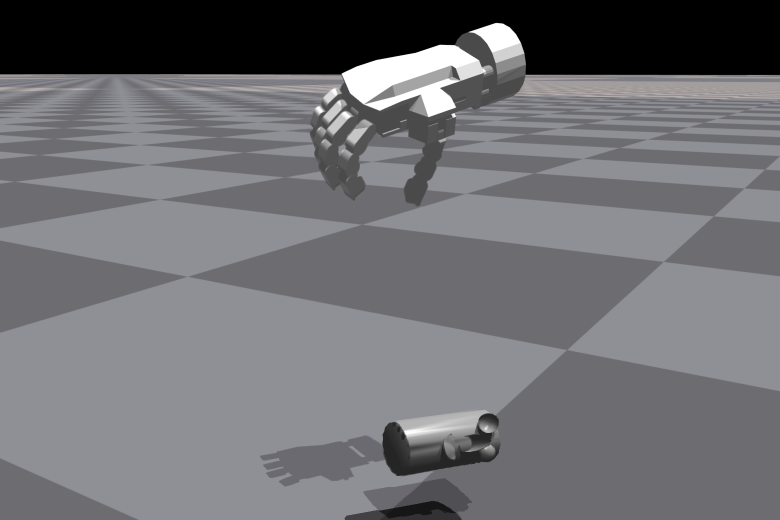} &
        \includegraphics[width=0.156\textwidth]{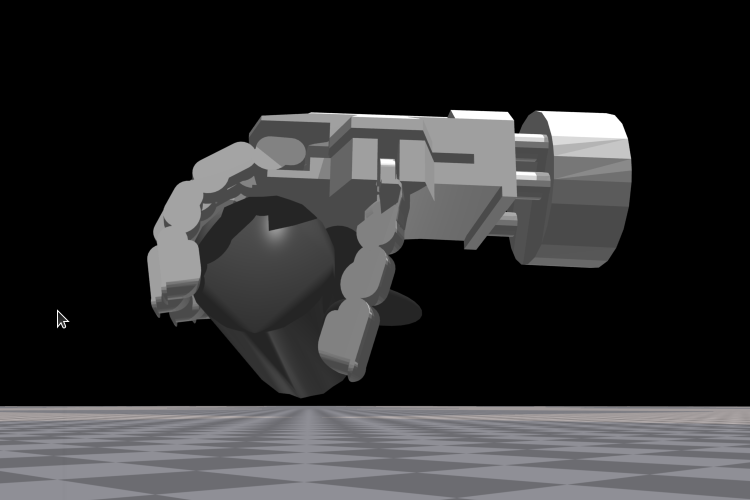} &
        \includegraphics[width=0.156\textwidth]{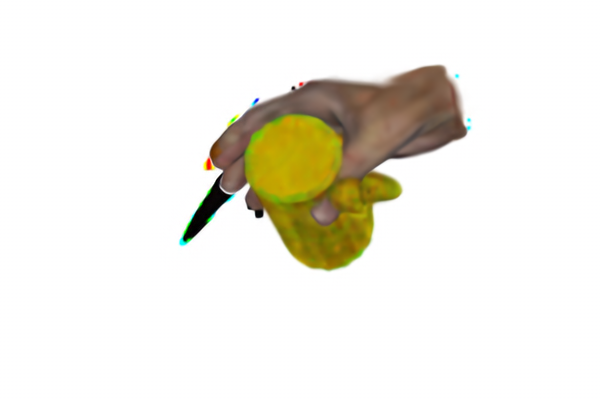} &
        \includegraphics[width=0.156\textwidth]{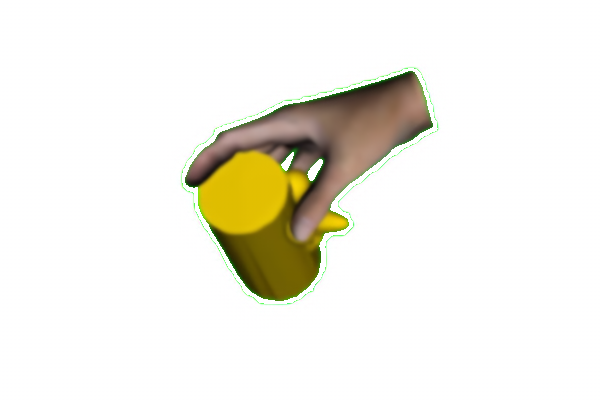} \\
        \includegraphics[width=0.156\textwidth]{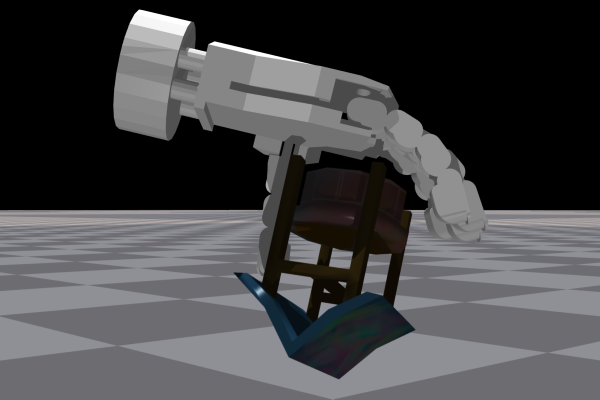} &
        \includegraphics[width=0.156\textwidth]{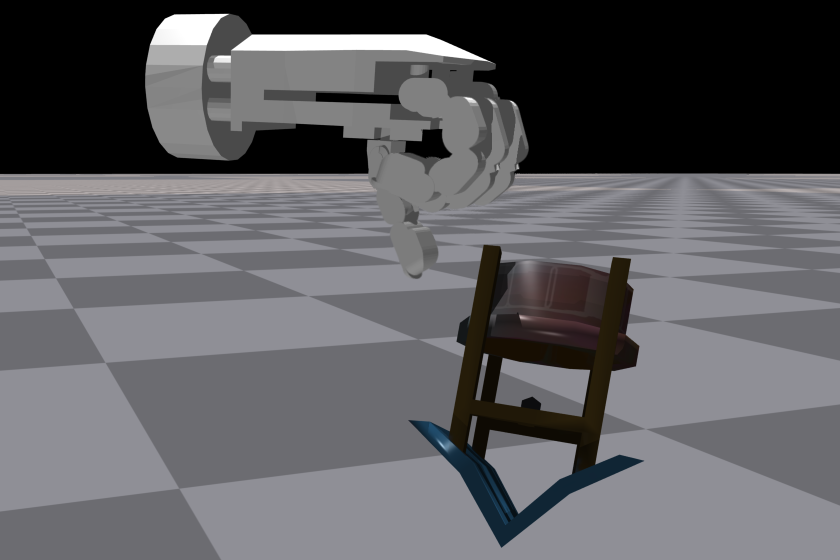} &
        \includegraphics[width=0.156\textwidth]{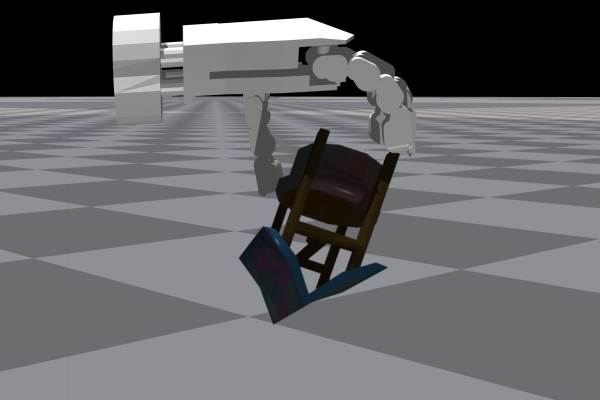} &
        \includegraphics[width=0.156\textwidth]{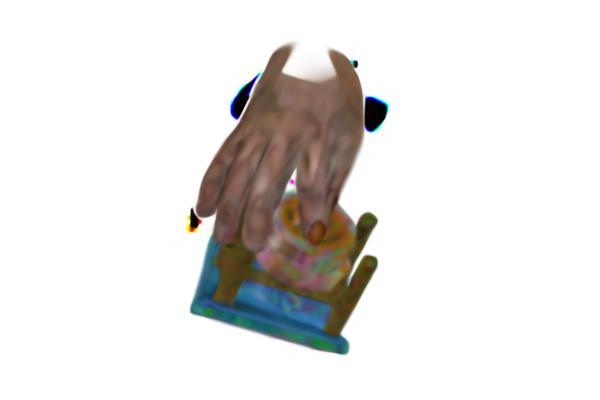} &
        \includegraphics[width=0.156\textwidth]{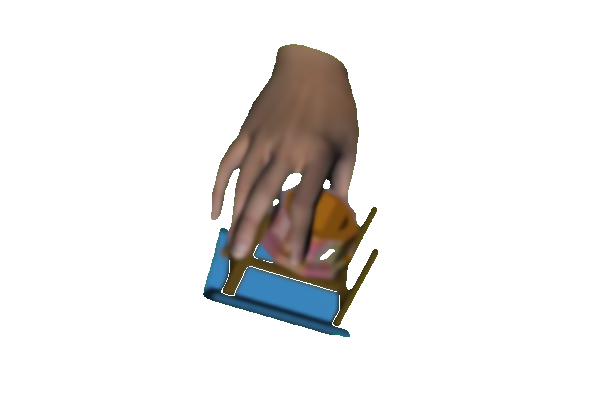} \\
      \end{tabular}
      \vspace{-2mm}
      \caption{\ourmethod dataset}
      \vspace{4mm}
      \label{fig:visual_graspxl}
    \end{subfigure}
    
    \begin{subfigure}[b]{\textwidth}
      \centering
      \renewcommand{\arraystretch}{1.2}
      \begin{tabular}{cccc}
        GANHand \cite{corona2020ganhand} & Affordance Diffusion \cite{ye2023affordancediffusion} & Ours (Simulation) & Ours (3DGS) \\
        \includegraphics[width=0.156\textwidth]{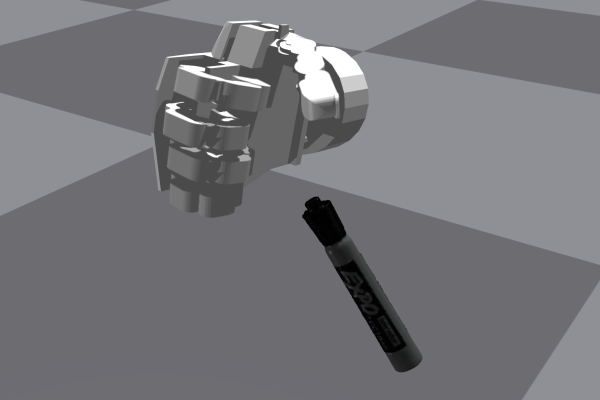} &
        \includegraphics[width=0.156\textwidth]{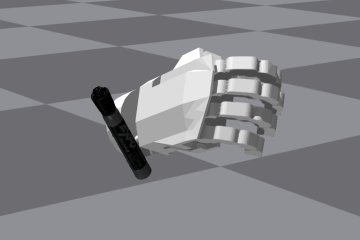} &
        \includegraphics[width=0.156\textwidth]{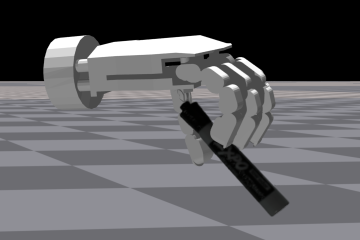} &
        \includegraphics[width=0.156\textwidth]{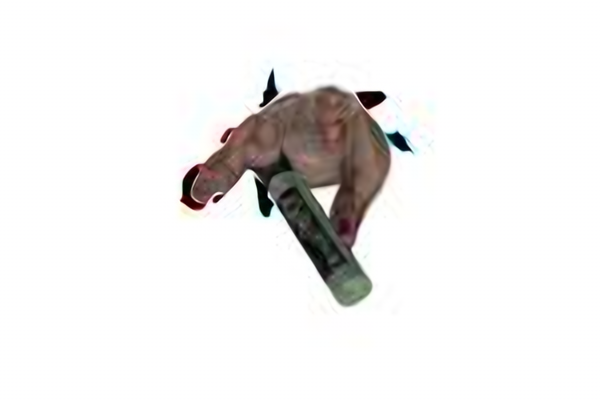} \\
        \includegraphics[width=0.156\textwidth]{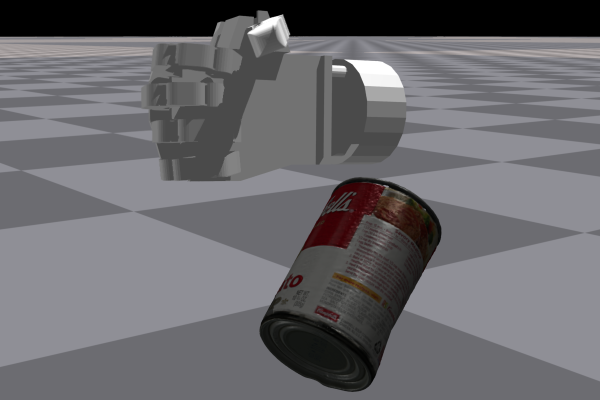} &
        \includegraphics[width=0.156\textwidth]{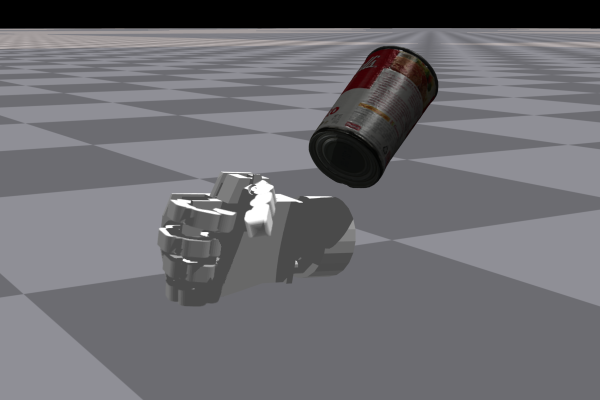} &
        \includegraphics[width=0.156\textwidth]{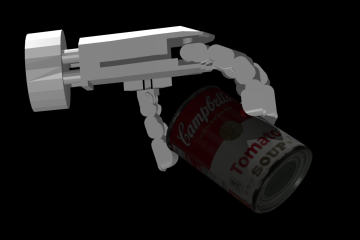} &
        \includegraphics[width=0.156\textwidth]{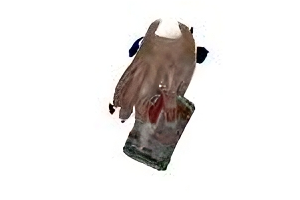} \\
      \end{tabular}
      \vspace{-2mm}
      \caption{YCB dataset}
      \label{fig:ycb}
    \end{subfigure}
    
    \caption{Comparison of methods on the \ourmethod (top) and YCB (bottom) datasets. Each row represents a different object. For each of the compared methods, as well as for \ourmethod, we present the results in the simulated environment. Additionally, for our method, we also show qualitative image synthesis of our 3DGS model during training against the relative ground truth.
    }
    \vspace{-1.5em}
    \label{fig:qualitative_results}
\end{figure*}

\noindent\textit{Datasets.} 
Novel view synthesis at training time requires a dataset with objects seen from multiple viewpoints. Likewise we need a 3DGS model for the hand as described in Sec.~\ref{sec:3dgs_phot_loss}. Current datasets for object grasping such as ~\cite{liu2022hoi4d} do not provide this information, so the need to create the \ourmethod dataset.
Our dataset includes a reference pose to evaluate performance using the Mean Joint Error, which measures the average discrepancy between the predicted MANO joints and their ground truth counterparts. 
We compare methods based on rotation and translation errors at the base of the hand relative to the ground truth.
These evaluations are performed on the test partition, which is obtained through an 80\%-20\% split ratio. 
Since our goal is to achieve successful grasping, we further assess performance using a physics simulator. 
For the test set, we measure the percentage of objects that can be successfully grasped using a QB SoftHand~\cite{catalano2014softhandQB} model mapped onto the MANO model.
We align both hands relative to their wrists and apply a 5\textdegree{} rotation along the y-axis to compensate for the original MANO's skewness. Our simplified SoftHand model accurately simulates most of the QBHand’s behavior, including its underactuated joints. A grasp is considered successful when at least two fingers contact the object.

The quantitative results of these experiments are presented in Tab.~\ref{tab:comparison_results}.
To evaluate generalization capabilities, we also conduct a zero-shot test on the YCB dataset~\cite{Calli2015ycb} with randomly oriented objects. 
Since no reference hand poses are provided in YCB, we neither train nor fine-tune the model on YCB objects. 
Instead, we directly apply the network trained on our dataset to YCB objects, assessing its generalization performance. 
Due to the lack of ground truth hand poses in YCB, we report only grasping accuracy in Tab.~\ref{tab:comparison_results}.

\noindent\textit{Comparison methods.}
We compare \ourmethod against GANHand and AffordanceDiffusion, two methods designed to predict a grasping pose from an RGB image.
GANHand \cite{corona2020ganhand} directly estimates the pose in 3D, whereas AffordanceDiffusion \cite{ye2023affordancediffusion} utilizes diffusion models to generate an image of the grasping interaction.
To ensure a fair comparison, we retrained both methods on the \ourmethod dataset using their original repositories. However, during the training of GANHand, we observed catastrophic failures. To stabilize the learning process,  we enabled the forward kinematic losses present in the original GANHand implementation~\cite{corona2020ganhand}.

\vspace{1mm}
\noindent\textit{\ourmethod implementation details.}
Our method is implemented in PyTorch and trained for 1.5K iterations ($\sim$ 13 hours) using a NVIDIA RTX 4500. 
The translation normalization is based on the objects' size.
We employ the AdamW optimizer~\cite{loshchilov2019adamw} with a weight decay of $10^{-2}$ and a learning rate of $10^{-5}$.
A step learning rate scheduler is used, decaying every 1,900 iterations with a decay factor of 0.7.
For 3D Gaussian Splatting (3DGS) rendering, we utilize gSplat~\cite{ye2024gsplat}.
The grasping simulations are implemented using IsaacGym~\cite{makoviychuk2021isaacgym}.

\subsection{Experimental evaluation}
\label{sec:discusssion}

Tab.~\ref{tab:comparison_results} presents the results on \ourmethod  and YCB datasets. No method was finetuned for YCB so evaluation is zero-shot for this dataset.
The first part of the table shows that \ourmethod achieves state-of-the-art performance on both datasets, outperforming prior works.
We achieve a 64\% improvement in grasp accuracy on \ourmethod and 41\% on YCB.
The small difference in translation and rotation errors across methods, compared to grasp accuracy, indicates that minor mispredictions can significantly impact grasp success.
This effect is evident in the inverted well toy from \ourmethod (Fig.~\ref{fig:qualitative_results}), where our method successfully grasps the object between the thumb and index finger despite the minimal contact surface.

The zero-shot test on YCB  further demonstrates \ourmethod's ability to generalize to out-of-distribution objects.
In the same situation, GANHand has much lower performance due to higher translation errors that impair successful grasps.
Fig.~\ref{fig:qualitative_results} shows that while Affordance Diffusion can estimate reliably the multi-fingered hands joint poses, but it severely struggles to estimate translation.
This problem is due to the method reliance to an off-the-shelf hand pose estimator that is not able to generalise to different scenarios. This affects the estimation of the distance of the hand from the camera, and in turn leads to wrong translation estimates.
This also affects accuracy, in many cases the hand collides with the object, which we consider a grasping failure.
In our experiments, we found that despite the mapping the hand models differ slightly. For example, QBHand has a flat palm, while MANO's is slightly convex. Even a small correction of $2 \text{ mm}$ can cause a 1–2\% performance drop.

The second part of Tab.~\ref{tab:comparison_results} presents the ablation studies. 
Lowering image resolution (224$\times$224) or stride marginally reduces accuracy on the training set but drastically impacts performance on new objects like those in YCB, likely due to the loss of fine detail reasoning. 
However, reducing stride or resolution significantly speeds up inference (134 ms), highlighting a performance-efficiency trade-off. Note that even with the fastest inference, \ourmethod still tops all competing approaches in all the metrics.
Additionally, removing direct hand parameter optimization losses ($\mathcal{L}_{pose}$ and $\mathcal{L}_{transl}$) has minimal impact on accuracy but increases convergence time.

\section{Conclusions}

In this work, we propose a novel dexterous grasping method, \ourmethod, that overcomes the limitations of existing approaches by leveraging 3D information while being trained solely on RGB images. 
Our method introduces a new paradigm for predicting successful grasps by synthesizing physically plausible hand-object images, enabling us to regress the necessary hand joints for a successful grasp. 
By utilizing 3D Gaussian Splatting to generate high-fidelity scene representations, we combine 3D consistency with the simplicity of RGB data, eliminating the need for complex 3D scans.
Experiments on both synthetic and real-world grasping datasets validate the effectiveness of \ourmethod, demonstrating significantly improved grasp success rates compared to baseline methods. 
\ourmethod provides a scalable and efficient solution for large-scale dexterous robotic grasping, opening new research directions in robotic manipulation. 

\bibliographystyle{IEEEtran}
\bibliography{main}

\end{document}